# Sentiment Analysis of Persian Language: Review of Algorithms, Approaches and Datasets

Ali Nazarizadeh[1] and Touraj Banirostam[2] and Minoo Sayyadpour[3]

[1] Computer Eng. Dep., Faculty of Technology and Engineering, Islamic Azad University, Central Tehran Branch, Tehran, Iran
`computer.engineer.as@gmail.com`

[2] Computer Eng. Dep., Faculty of Technology and Engineering, Islamic Azad University, Central Tehran Branch, Tehran, Iran
`banirostam@iauctb.ac.ir`

[3] Kharazmi University, Tehran, Iran
`std_minoosayyadpour@khu.ac.ir`

**Abstract.** Sentiment analysis aims to extract people's emotions and opinion from their comments on the web. It widely used in businesses to detect sentiment in social data, gauge brand reputation, and understand customers. Most of articles in this area have concentrated on the English language whereas there are limited resources for Persian language. In this review paper, recent published articles between 2018 and 2022 in sentiment analysis in Persian Language have been collected and their methods, approach and dataset will be explained and analyzed. Almost all the methods used to solve sentiment analysis are machine learning and deep learning. The purpose of this paper is to examine 40 different approach sentiment analysis in the Persian Language, analysis datasets along with the accuracy of the algorithms applied to them and also review strengths and weaknesses of each. Among all the methods, transformers such as BERT and RNN Neural Networks such as LSTM and Bi-LSTM have achieved higher accuracy in the sentiment analysis. In addition to the methods and approaches, the datasets reviewed are listed between 2018 and 2022 and information about each dataset and its details are provided.

## 1 Introduction

Sentiment Analysis is one of the most important topics in artificial intelligence and its main purpose is to understand emotions according to the input data. Depending on the type of input data such as video, image, text and voice, we can use machine learning algorithms and teach the machine to understand emotions. Thousands of websites and blogs updates and modifies by Persian users around the world that contains millions of Persian contexts. This range of application requires a comprehensive structured framework to extract beneficial information for helping enterprises to enhance their business and initiate a customer-centric management process by producing effective recommender systems. This language is an Indo-European language which spoken by over 110 million people around the world and is an official language in Iran, Tajikistan, and Afghanistan [1].

Despite numerous capabilities of sentiment analysis in Persian language, there are limitation in this area such as: lack of researches, reviews, datasets and libraries. The main concern of sentiment analysis based on Persian language are inability to perform well in different domains, inadequate accuracy and performance in sentiment analysis based on insufficient labeled data, incapability to deal with complex sentences that require more than sentiment words and simple analyzing. Algorithms in this field should be compared by multiple criteria such as classification type, accuracy, method and dataset.

There are a few technical surveys and review papers between 2018 and 2022 [2,3,4] that do not concentrate on the multi criteria comparing in sentiment analysis algorithms in Persian language. The main aim of this research is to survey in different sentiment analysis algorithms in Persian language to comprehend the diversity of approaches in this area which have been recently presented between 2018 and 2022. The key approaches of sentiment analysis that have been focused in selected studies consist of machine learning and deep learning approaches. We present algorithms, approaches and datasets review and overview opportunities of sentiment analysis in Persian language.

The main commitments of this study are highlighted as follows:

- Designing a technical taxonomy to classify sentiment analysis approaches in Persian language between 2018 to 2022.
- Presenting a discussion of the key challenges for different approaches in sentiment analysis in Persian language.

- Highlighting the future research challenges and open issues in sentiment analysis in Persian language.

The organization of this review is considered as follows: Section 2 describes the background, and Section 3 explains the related reviews and surveys in Sentiment Analysis in Persian language. Section 4 provides the research methodology, and Section 5 presents a taxonomy of the selected approaches and discusses challenges, open issues and future trends. Finally, Section 6 demonstrates the conclusion along with the paper restrictions.

## 2  Backgrounds

In this section, a brief definition of sentiment analysis is presented. At first, an outline of sentiment analysis is explained. Then, sentiment analysis in Persian language is described. Finally, the important metrics that are used in this subject like classification type are defined.

### 2.1  sentiment analysis

The term sentiment analysis was defined by Bo Pang, Lillian Lee [5] in early 2000's. Sentiment analysis, also referred to as opinion mining, is an approach to natural language processing (NLP) that identifies the emotional tone behind a body of text. Sentiment analysis systems help organizations gather insights from unorganized and unstructured text that derive from online sources such as emails, blog posts, support tickets, web chats, social media channels, forums and comments. Algorithms replace manual data processing by implementing rule-based, automatic or hybrid methods. Rule-based systems perform sentiment analysis based on predefined, lexicon-based rules while automatic systems learn from data with machine learning techniques. A hybrid sentiment analysis combines both approaches [6].

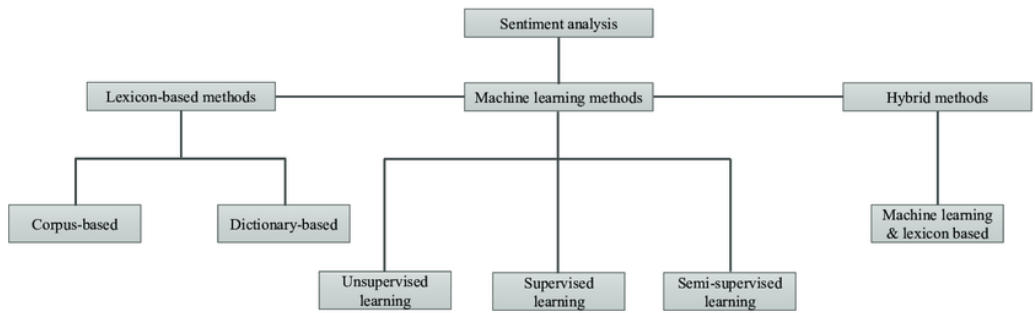

**Fig. 1.** Categories of sentiment analysis [6].

### 2.2  Types of sentiment analysis

Research on Sentiment Analysis is divided into four categories based on the level of decision making. (1) Word-level Analysis (or phrase-level analysis), (2) Sentence-level analysis, (3) Text-level analysis, (4) Feature-level analysis (or aspect-level analysis).

In word-level analysis, the focus is on the polarity of the words and whether they are positive or negative. At this area, all words have polarity and degree of being positive or negative and the main method is to use the Lexicon of emotions, and based on this, classification of emotions is done. In sentence-level analysis, the polarity of sentences is determined, and instead of identifying the polarity of words, the polarity of sentences in the text is determined. The task of text-level analysis is to classify text, news, user comments, movie reviews, or more of the same. The goal at this level is to identify the polarity of each sentence and classify it into two or more classes. Based on the dataset in sentiment analysis, emotions are classified into two classes (negative, positive), three classes (negative, neutral, positive), or five classes (angry, a bit negative, neutral, a bit positive, positive). Therefore, the polarity of short or long texts that contain several sentences can be determined. The aspect-level analysis is slightly different from previous methods. At this level, polarity is defined based on one or more features in the text. For example, look at this phrase: "The Phone's Camera is Great but Expensive". In this phrase, emotions can be examined based on the two characteristics of "Camera" and "Price" that Feeling for the "Phone's Camera" is positive but Feeling for the "Price" is negative.

## 2.3 Metrics definition

This section defines the metrics used to evaluate the proposed sentiment analysis approaches. These metrics are based on confusion matrix. A confusion matrix contains information about actual and predicted classifications done by a classification system, in this case sentiment analysis approaches.

- *Accuracy:* In sentiment analysis, accuracy as a score derived by the ratio of correct predictions to total predictions made by a classification system.
- *F1-score:* In statistical analysis, the F1 score (also F-score or F-measure) is a measure of a test's accuracy. It is calculated from the precision and recall of the test, where the precision is the number of correctly identified positive results divided by the number of all positive results, including those not identified correctly, and the recall is the number of correctly identified positive results divided by the number of all samples that should have been identified as positive. F1 score can also be described as the harmonic mean or weighted average of precision and recall. These formulas for binary classification are shown in Figure (2).

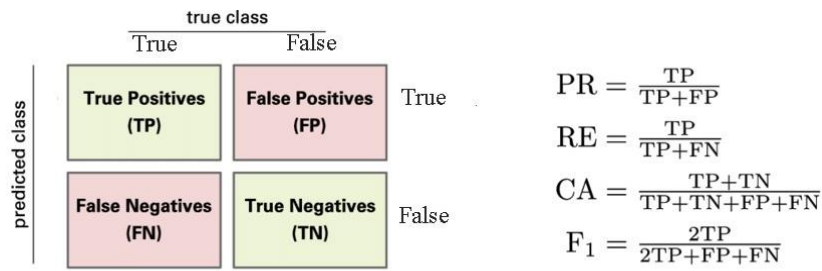

**Fig. 2.** Confusion matrix with the formulas of precision, recall, accuracy and f1-score

## 3 Research selection methods

This section provides a review based on SLR method as a research study assessment for classifying sentiment analysis in Persian language. Contrary to an unstructured review process, the SLR method was applied to reduce bias and to follow a precise sequence of methodological phases to research literature. An SLR depends on the well-defined review protocol to extract, evaluate, and document results as depicted in Figure (3). Considering the alternatives and other synonyms of the key essential components, the subsequent exploration string was defined:

- ("Persian") AND ("sentiment analysis" OR "opinion mining")
- ("sentiment analysis" OR "opinion mining") AND ("in Persian" OR "in Farsi")

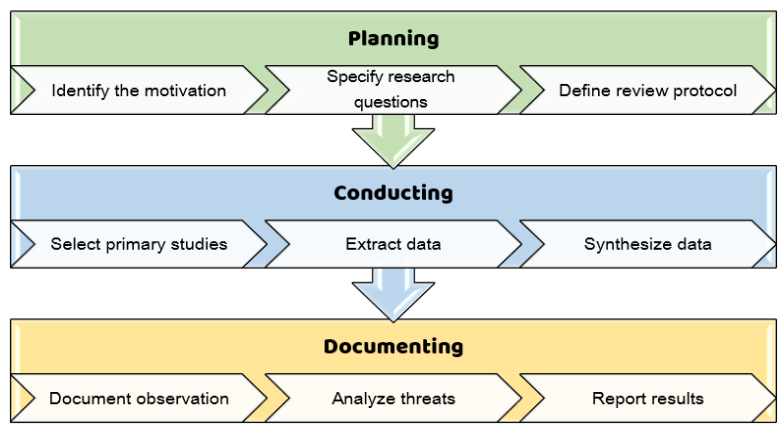

**Fig. 3.** An overview of our research methodology.

This SLR paper presents inclusive answers to the following Analytical Questions (AQ) regarding the aims of this review:

AQ1: Which methods in machine learning are considered for Sentiment analysis in Persian language?
AQ2: Which classifiers are categorized in Sentiment Analysis in Persian language?
AQ3: What datasets are used for evaluating Sentiment analysis in Persian language?
AQ4: What are the evaluation factors usually applied in Sentiment analysis in Persian language?
AQ5: What are the innovations used for increasing accuracy in Sentiment analysis in Persian language?
AQ6: What are the future researches directions and open perspectives of Sentiment analysis in Persian language?

Figure (4) shows the distribution of the research studies completed by the leading scientific publishers regarding the article citations and review method, including IEEE, ScienceDirect, Springer, ACM, Google Scholar. Figure (5) shows the distribution of the research studies by year they published.

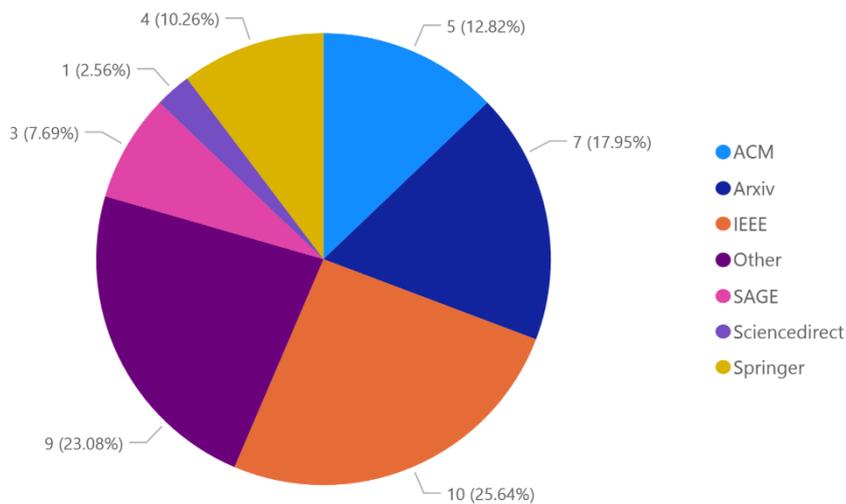

**Fig. 4.** Distribution of research papers by publisher.

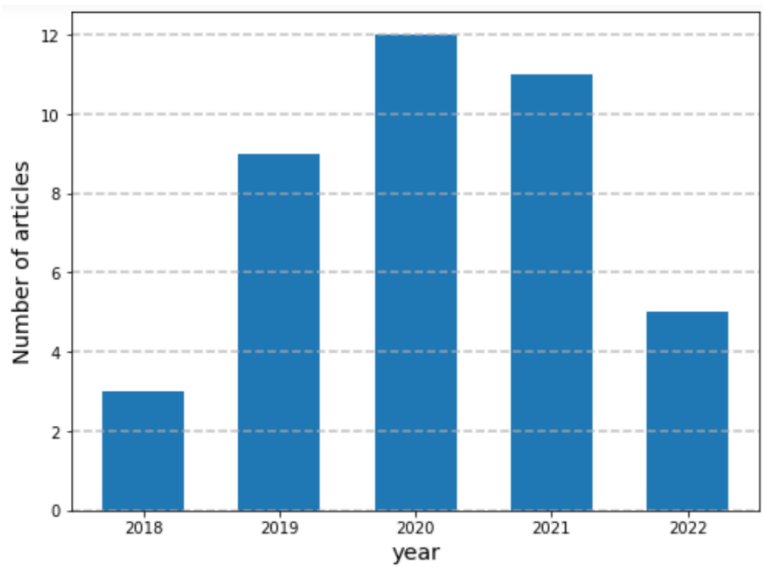

**Fig. 5.** Distribution of research papers by published year.

After providing the analytical questions, the inclusion/exclusion criteria for the ultimate research selection were applied. Regarding the number of published papers, we just analyze the journal articles and conference papers between 2018 and 2022 for Persian sentiment analysis.

Finally, 40 papers were selected for analyzing and answering the mentioned analytical questions which are presented in detail in Section 4.

Figure (6) shows the selection principles and evaluation flowchart designed for the studies. The exclusion phase consists of ignoring short papers, book chapters, and low-quality studies (published in predatory journals) that did not present any scientific discussion and technical information. For mapping the final selected studies, the inclusion principles are considered as follows:

- The studies available online between 2018 and 2022.
- The studies in Persian sentiment analysis topics.
- The studies with machine learning and deep learning methods in sentiment analysis.

For mapping the final selected studies, the exclusion principles are considered as follows:

- The studies presenting survey papers.
- The studies not indexed in ISI.
- The studies not written in English language.

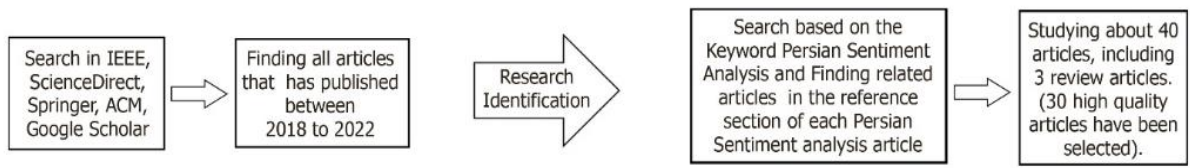

**Fig. 6.** The selection criteria and evaluation chart of research studies.

## 4 Different approaches in sentiment analysis

The related surveys and systematic reviews in Sentiment analysis in Persian language is discussed in this section to disclose the lack of comprehensive reviews and indicate the benefits and weaknesses of applied approaches systematically and taxonomically.

In [7], A Hybrid Persian Sentiment Analysis Framework was developed. This method used the Integrating Dependency Grammar Based Rules and Deep Neural Networks. The focus of this article is on new rules based on the dependency for the analysis of Persian emotions hierarchical relationship between keywords, words, word order, polarities (positive and negative comments) in long sentences, and constraints based on the frequency of synchronous words. The method used in this article has a higher accuracy of 18-16% compared to traditional machine learning methods and 6-9% compared to deep learning methods.

In [8], feature extraction in the Persian language was accomplished using the hybrid feature extraction based on convolutional neural networks. Three main steps have been considered for their method: The first step involves preparing the sentence for the input matrix, which is done in two levels: word level and character. At the word level, each word in each sentence is given to the word2vec algorithm. At the character level, for each character in each sentence, the proposed method calculates a numerical vector and creates a matrix. Then the feature extraction part which includes a convolutional neural network is executed. The matrices generated in the previous levels are given to the CNN for each sentence to embed each sentence. Therefore, using word2vec and CNN to extract the properties in the last step, the generated vectors are given to the two-way short-term memory network to classify emotions, which has not been used in any of the previous methods.

In [9], the authors proposed a novel approach to sentiment analysis in Persian using discourse and external semantic information. The proposed method in this paper consists of two parts: the first method is based on a classifier combination and the second approach is based on deep neural networks. Both methods use information from the local discourse and the external knowledge base. And also cover several linguistic issues such as negation and accent, which involve different levels, namely the level of word, text, sentence, phrase, and document.

In [10], a Novel Deep Learning Models Trained Over Proposed Augmented Persian Sentiment Corpus was used. In this paper, the dataset is modified using another version of the SentiPers package that contains additional data and amplified by using the data noise technique. Then data translation applied that intends to replace only a few words with their synonyms. According to this idea, 38% of each sentence is selected for replacement. In this approach, three methods of classification are used: Naive Bayesian, Gradient descent, and Support Vector Machine (SVM). Then we compare these methods to deep learning methods. Deep learning algorithms have reached the best accuracy. The method used in this article is Bi-LSTM neural networks.

In [11], a translation-based method that combines several polar languages for use in English has proposed. And it used to produce a polarity dictionary in the target language (Persian) called Sentifars. To build Sentifars been used source NRC Emotion Lexicon, SentiWordNet 3.0, SenticNet 3.0, and Liu's Polarity Lexicon. The problem in this article is the construction of a glossary for non-English and non-source languages that are used in Persian. For proposed method, a large list of Persian terms (PT) to pass to the next step (labeling) is required. English-Persian-English dictionaries, as well as English WordNet, have been used for this purpose. In the next step, all documents are labeled manually. As mentioned, four polarity dictionaries have been used for feature extraction. And from these dictionaries, the most important features that are important for analyzing emotions have been extracted. For classification, the logistic classifier is used which receives the features of the previous step as input and performs the classification work.

In [12], an ensemble-based classification approach proposed for Persian Sentiment Analysis using machine learning and deep learning algorithms. In this article initially, preprocessing and normalization sections were done on text data. Then the classification was done by using three machine learning algorithms (SVM + MLP + CNN). The accuracy of those algorithms in order is 68%, 74%, 78%, and the innovation used in this article is an ensemble-based classification (three classifications) with an accuracy of 79.68%.

In [13], two models of deep learning algorithms are used for sentiment analysis in the Persian language. These algorithms are autoencoder neural networks and CNN networks and have been compared to MLP neural networks where deep learning algorithms have achieved the best accuracy. MLP neural networks at 100 iterations have reached convergence and their accuracy is 78%. Autoencoder neural networks also have an input layer and subsequent layers have 1500, 512, 1500 neurons. In the end, there is also an output layer that the accuracy of this model is 80%. The next model is CNN neural networks which have three layers: input layer, hidden layer, and output layer. The hidden layers contain convolutional layers, pooling layers, and fully connected layers. In the end accuracy of the model is 82%.

[14] has focused on emotion analysis using deep learning algorithms which perform emotion analysis for cross-lingual training for low-resource languages excellently. In this research using transfer learning, the problem of lack of resources in non-English languages has been solved. And this proposed method is not only for Persian but also works well in English to analyze emotions. Proposed architecture for cross-lingual sentiment analysis includes a neural model for word embedding and a deep neural network for document classification. For each part of this framework, proposed using different architectures. This architecture is general enough to be used with any embedding or any deep classifier. Such a general framework makes the system flexible to be used for sentiment analysis of any low-resource language. Various classification algorithms have been used, such as CNN and LSTM, as well as a combination of the two, LSTM-CNN and CNN-LSTM, which have achieved good results.

M.B. Dastgheib in [15] proposed a hybrid method by a combination of structural correspondence learning (SCL) and convolutional neural network (CNN) for sentiment analysis. The task of the SCL method is feature extraction and it selects the most effective pivotal features that ultimately aim to improve classification performance using deep learning. SCL algorithm uses labeled data (xt, yt). In the first step of the algorithm, m central features are selected from the main source data (pi represents the first pivot property). In this article, L(p,y) is a real loss function for classifying binary problems. Using the SCL algorithm, the original feature space is reduced to the h dimensional space of axial features. The next step is to train CNN neural networks using these features. Each sentence is represented using axial properties in the h dimensional space. CNN neural networks have one input layer, six hidden layers, and one softmax output layer.

[16] presented a model for Tehran Stock Exchange prediction using Sentiment Analysis of online textual opinions. In fact, in this article, the effect of social media data in predicting the variables of the Tehran Stock Exchange has been studied for the first time. The method introduced in this article is a combination of vocabulary-based and learning methods. Since there is no glossary in Stock Exchange Prediction, a new glossary has been created. And by using combination and machine learning algorithms, emotion analysis has been used to predict the Tehran Stock Exchange. After collected user's comments on the "www.sahamyab.com/stocktwits" website and labeling, comments were classified using Naive Bayesian, Decision Tree, Support Vector Machine, and Bagging J48. Among them, Bagging J48 has reached the best accuracy.

In [17], exploited BERT to improve Aspect-Based Sentiment Analysis (ABSA) performance on the Persian language was used. The present study aimed to improve ABSA in the Persian pars-ABSA dataset and this research has the potential to use a pre-trained BERT model. Pars-BERT pre-trained model in combination with NLI-M auxiliary sentence shows the best performance and increases the accuracy to 91% for an ABSA task on the Pars-ABSA dataset which is more than 5.5% (absolute) higher than the previous best model.

In [18], deep learning and active learning methods have been used to analyze Persian emotions. In this article, data is transferred to a meaning-based Latent space (LDA) and samples are extracted and selected in this space. The LDA is a three-tier Bayesian hierarchical model that summarizes these two levels: Clustering each word in a cluster (each topic is a possible distribution of words). And Document clustering is a combination of extracted topics (use bayesian inference to assign each document to a probability distribution in topics). In this article, active learning strategies in LSTM-based architecture, as well as an architecture based on CNNs, are reviewed.

In [19], a combined deep learning model is used to analyze emotions in Persian. Local features are extracted by CNN neural networks and long-term dependencies are learned by LSTM. The main method after creating a Persian dataset, the word2vec algorithm was used to embed each sentence. CNN extracts local features that different core sizes are used and a modified linear unit (Relu) is also used to activate the output of the CNN layer. A pooling layer is used to create a high-level feature and a CNN neural network is used to create a low-level feature. These low-level features are given to an LSTM neural network for emotion classification. LSTM can learn long dependencies between words and finally, a fully connected layer is used to categorize each comment.

In [20], a new model based on aspect-based emotion analysis (Pars-ABSA) is introduced. Pars-ABSA dataset contains 10,000 user comments in social media (5,114 negative comments, 3,061 positive comments and 1,827 neutral comments). Finally, a Bi-LSTM neural network has been used to classify emotions.

M.E Basiri in [21], a comprehensive study of the dictionary-based method has been done, and then two new sources have been introduced to solve the problem of lack of Persian resources in the field of emotion analysis that Includes a carefully labeled glossary of emotional words (perLex) and a new handmade dataset with approximately 18,888 rated documents (PerView). In addition, a new hybrid method using both machine learning and dictionary-based methods has been proposed where the words PerLex are used to teach the machine learning algorithm. In the dictionary-based approach, the words of the words and the label of the emotions related to them are used to determine the overall feeling of a sentence or a document. Compared to machine learning methods, this method has several advantages such as strength, range independence, ease of implementation and the ability to improve using different sources of knowledge. After initial processing and input analysis, lexicon terms, bigrams were used in the feature extraction step.

In [22], a new combined method for ranking prediction in Persian was proposed. Experimental results in a large dataset containing 16,000 Iranian customers show that this proposed system achieves higher performance than the Naive Bayes algorithm and a dictionary-based method. The main method that has been improved in this study is the dictionary-based method for exploring surveys in Persian. In addition, the results suggest that this proposed method may also be used successfully to detect polarity.

In [23], sentiment analysis in Persian Language based on aspects has been researched, but due to the lack of labels on comments, clustering methods have been used. Using unlabeled data, a Word2vec model is trained using neural networks. The proposed solution for extracting aspects is to use a combination of rule-based methods and neural networks using the Word2vec model. In this study, first, the candidate aspects are extracted using rule-based methods, then the most candidate aspects are selected and the final aspects are extracted using clustering methods and cosine similarity criteria. In the method proposed in this research, 68 of the most volunteer aspects have been selected, which have been selected in the feature extraction stage. These 68 aspects and their related Word2vec vectors are given as K-Means input. The main advantage of K-Means output in this section is that it can show us the distribution of the main aspects in the vector space with appropriate accuracy.

In [24], an ontology-based emotional dictionary is used to analyze emotions in Persian. In this article, first, the emotional dictionary is produced and labeled, then a graph-based method based on ontology is introduced, and finally, the KNN algorithm is used for classification. In the first two steps, a semi-supervised method was employed for annotating a set of FarsNet synsets. These annotated synsets can then be used as a training set for classifying the rest of the synsets. This paper explains how to construct a graph representing the FarsNet ontology. Each synset in the ontology is mapped to a node in the graph. Since ontology relations are defined between two synsets, each relation can be mapped to an edge in the graph. In this approach, each synset in FarsNet ontology must become a document. A frequency-based method (TF-IDF) was used to extract the feature.

In [25], a Sentiment Analysis Corpus for Persian called SentiPrers is provided. This collection contains 26,000 annotations in Persian. One of the important features of SentiPers is the inclusion of formal (written) and informal (verbal) sentences. The set includes annotations at all three levels, including document level, sentence level, and entity/aspect level. The data used to build SentiPrers is taken from the digikala website. Output tags are divided into 6 categories, which are displayed with the numbers -3 to +3, -3 means angry feeling about the product and +3 means the highest satisfaction about the product. The number zero also means a neutral state.

In [26], an improved evidence-based aggregation method for sentiment analysis is provided. In this study, a new method for aggregating scores is presented that uses both the maximum and the second possible classes to predict the final score. In the proposed method, each text is considered as a set of sentences, each with its orientation and score, and the probability that each sentence belongs to different classes on a five-star scale, using a system based on the dictionary is calculated. These probabilities are then used to identify emotions at the document level. In this study, a dictionary-based method has been used to analyze emotions at the sentence level, and one of the best emotion dictionaries called SentiStrength dictionary has been used.

In [27], a novel approach has been proposed to identify the polarity of the Persian sentences. This approach has been proposed to extract POS tag of the Persian sentence, then PerSent lexicon (Persian lexicon) has been used to assign polarity to extracted features and SVM, MLP and Naïve Bayes classifiers has been used to evaluate the performance of the approach. The SVM received better results in comparison with Naive Bayes and MLP. Persian movie review dataset has been used to evaluate the performance of the proposed approach. The weakness of proposed approach for Persian sentence polarity detection is not able to handle sarcasm in the sentence. Further study is required to detect ironic and sarcasm in the sentence to improve the performance of the approach.

In [28], a method for adapting the domain from formal to colloquial in sentiment classification has been presented. This method uses two approaches, adversarial training and weak supervision, and only needs a few shots of labeled data. In the first stage, authors labeled a crawled dataset (containing colloquial and formal sentences) with weakly supervised sentiment tags using a sentiment vocabulary network. Then they fine-tuned a pre-trained model with adversarial training on this weak dataset to generate domain-independent representations. In the last stage, they trained the above fine-tuned neural network with just 50 samples of data (formal domain) and tested it on colloquial. Dataset of formal and colloquial sentences (crawled from Digikala) include 20,000 sentences taken from expert review texts and 20,000 sentences taken from user comments of products has been used for this paper. Dataset includes the opinions of users and experts of Digikala which are human-labeled (an integer between -2 to +2). Also, HesNegar Persian sentiment vocabulary network which includes 100062 unique words with the weight of each word in the positive and negative classes. Experimental results show that proposed method with only 50 formal data has raised the F1 to 80%.

In [29], a sentiment analysis model is trained using cross-lingual word embeddings and in order to decrease the dependency to training data an Active Learning algorithm is applied to the sentiment model. Cross-lingual model trains a model by using a rich-resource language like English as a source language and apply it to a low-resource language. The dataset used as the training data is Amazon Food Reviews. Persian Snapp Food reviews dataset is used as the test dataset. Persian dataset is contained of 6,456 reviews both negative and positive. As the English dataset, we have 90,000 reviews out of 104,536 as the train data set and 14,536 reviews as the test dataset. The English Model accuracy without Active Learning data was about 87%. By applying Active Learning to this model, the accuracy was increased significantly to about 85%. In the second level about 920 samples were chosen and this process has been continued until the 7th round which about 1400 samples were chosen. Proposed model accuracy was increased to about 90%.

In [30], over 12000 Persian tweets including the stock market keyword have been crawled from twitter. They are labeled manually in three different categories of positive, neutral and negative. Then a pre-trained ParsBERT model has been fine-tuned on these data. proposed model in this research is evaluated on the test dataset and compared to its counterpart, lexicon-based method using Polyglot as its lexicon. Accuracy of 82% has been achieved by proposed model surpassing its lexicon-based contender. Data were collected from January 1, 2020 to December 1, 2020 when the stock market experienced both significant growth of index and its sharp decline and subsequently were stored in a file formatted as excel. A total of 12055 tweets were collected, none of which contains missing values. Twint2 as a python package was used for extracting Persian tweets. The best value for the precision criterion was for the positive class with 89%. For recall and f-measure its best values were for the positive class by 87.4% and 88.2% respectively. On average, for the three classes, an accuracy of 82.04% has been achieved for the precision, 82.6% for the recall, and 81.83% for the f-measure.

In [31], a multimodal deep learning was presented for modelling relationship between the text and image modalities. the proposed model consisted of two parallel branches for text and image modalities. The text branch utilizes a bi-GRU layer to consider long sequential dependencies in the text. This branch also utilizes a word embedding layer to convert the text into a numerical dense representation. The image branch utilizes three two-dimensional CNN layers to extract meaningful features from the images. The outputs of the two branches are finally concatenated and sent to a fully connected layer for the final sentiment classification. In order to evaluate the proposed method, a multimodal dataset of Persian comment and their corresponding image, MPerInst, was manually collected and introduced in the current study. One limitation of this study is the relatively small size of the MPerInst dataset. Therefore, extending the MPerInst dataset and adding more fine-grained sentiment labels to it may be considered for the future research. Also, using pre-trained word embeddings and deep image networks may be another line for the future research. Proposed model accuracy was 92% and 92% for f1-score.

In [32], the effects of the sentiment lexicon, aggregation level, and aggregation method on the sentiment polarity and rating classification of Persian reviews are investigated. To this aim, a new sentiment aggregation method based on the cross-ratio operator is proposed. The results on four Persian review datasets show that the review-level aggregation can improve rating classification, although this approach does not have a positive impact on polarity classification. in this study, four datasets of customer reviews about different brands of cell phones were extracted from Digikala.com (Digikala 2017) in the period of July 2016 and February 2017. Proposed model accuracy was 81% ,69% for recall, 67% for precision and 68% for f1-score. The present study is limited to decision-level aggregation, while feature-level aggregation methods and their effects on the overall performance of the SA system should also be investigated. Another limitation may be the investigation of the effects of feature selection methods in general and latent-space models in particular.

In [33], authors used several News Agencies reports and reviews to identify potential events that had a possible association with the different sentimental trends in Twitter and used a deep learning model based on CNN-LSTM architecture for sentiment analysis and used three classifications. In this study, they analyzed the sentiments of 803278 Persian tweets concerning COVID-19 vaccines. Tweets retrieved mentioning the homegrown and imported vaccines including COVIran Barekat, Sinopharm, Pfizer/BioNTech, AstraZeneca/Oxford, and Moderna vaccines between April 1, 2021 and September 30, 2021. authors then grouped the retrieved tweets into two separated datasets: The Homegrown-vaccine dataset (for COVIran Barekat) and the Foreign-Vaccines dataset (for the rest of mentioned vaccines). One of this study's limitations is that collected tweets included just a short time of vaccine availability. Further work can focus on vaccine-related tweets after September, when most people were actively receiving vaccines. Furthermore, proposed method did not explore the attitude of Twitter users towards each vaccine separately.

In [34], an Attention-based LSTM, which has been shown to perform more effectively compared to the similar methods in sentiment analysis for the English language, has been used. The proposed method is evaluated on the two Persian "Taaghche" and "Filimo" datasets collected in this paper. The experiments on the two Persian datasets prove that utilizing informal vectors in sentiment analysis and applying the attention model improves the prediction accuracy of the DNN in the sentiment analysis of Persian texts. Both datasets are divided into two training and testing sets. The proposed algorithm is based on five consecutive steps: Text preprocessing, word embedding, LSTM network, Attention model, and the Cost function. Due to the effective differences between the two formal and informal contexts in Persian and the absence of word vectors for informal words, this paper provides a text corpus of informal Persian texts by applying Fasttext. "Filimo" dataset is the labeled dataset of Persian comments in three categories: positive, negative, and neutral and "Taaghche" dataset that includes "Taaghche" site users' comments about books on this website. The results indicate that the use of word vectors made from the informal text (80.3%) corpus is significantly more accurate than formal word vectors (76.4%). F1-score for informal texts was 73.4% and 67.8% for formal texts.

In [35], the insta-Text database is introduced, which contains about 9,000 comments from people following Instagram page of Hala Khorshid (HALA_KHORSHID). Comments are tagged in three categories with the help of 20 computer engineering students. Various preprocessing has been done on the database, including: checking duplication of comments, deleting English comments, normalizing the text, converting English numbers to Persian, correcting words, etc. Finally, word2vec has been trained on insta-Text and used to validate data.

In [36], the early version of Persian Sentiment Lexicon has been developed and the vocabulary of three areas of product and film, especially political news, has been added to it. For the news domain, hundreds of news articles have been collected from www.bbc.com/persian. For film's domain, one hundred articles from the site www.caffecinema.com have been used, and finally for product's domian, the contents of www.mobile.ir have been used, which contains the latest news of mobile technology. In this study, three annotators have identified the polarity of words with a number between -1 and +1, where -1 means the most negative sentiment polarity and +1 means the most positive sentiment polarity. 94% of the words are formal and the remaining 6% are informal.

In [37], a low-cost Persian Sentiment Analysis is performed without any Persian training data. A cross-lingual model between English and Persian is trained to generate aligned word embeddings that are used as the feature vectors in the sentiment model. Monolingual word embeddings used in cross-lingual approach are English FastText and Persian GloVe. VecMap method is used as the cross-lingual tool to make English and Persian word embeddings aligned in a supervised mode. Furthermore, a 5,000-word English-Persian bilingual dictionary is used as the supervision. Bilingual lexicon induction evaluation reveals that English and Persian are aligned properly in the joint space. The proposed Sentiment Analysis model is trained on an English dataset, and then is tested on Persian using aligned English-Persian word embeddings. The dataset used as the training data is Amazon Fine Food Reviews and Persian Snapp Food dataset is utilized as the test data. The model results show significant efficiency in the Sentiment Analysis task, though it does not use any Persian dataset in training procedure. The proposed cross-lingual Sentiment Analysis shows a good performance with F1-score of 78.16% on Persian test data.

In [39], aspect category detection (ACD) and aspect category polarity (ACP) are solved simultaneously and in one model, with the help of four deep learning models (CNN, LSTM, Bi-LSTM, GRU). In other studies, separate solutions have been proposed for these two tasks, but in this paper, a constant solution for both tasks have been developed in the same model.

In [40], a framework using TF-IDF and transition point to detect polarity in Persian movie reviews has been proposed. This approach has been evaluated using different classifiers such as SVM, Naive Bayes, MLP and CNN. The experimental results show the transition point is more effective in comparison with traditional feature such as TF-IDF.

In [41], PerSent Persian sentiment lexicon with 1000 idiomatic expressions has been developed. These words are useful not only for word recognition in Persian texts but also for accurate classification of Persian. With the help of three experts, the comments were labeled in two categories, positive and negative, with the accuracy of 0.1 decimal places (from -1 to +1). Very negative comments are labeled -1 and very positive comments are labeled +1. CNN neural networks were used for evaluation of classifying texts and these three datasets: Movie Reviews dataset, Persian VOA dataset and Amazon reviews dataset.

In [42], approximately 3 million comments have been extracted from the Digikala site, which has been tagged with two labels "I suggest" and "I do not suggest". However, a significant number of comments have not been tagged, which the pseudo-labeling technique has been used to tag unlabeled comments. As a result, 104,800 comments were positively tagged and 30,500 comments were negatively tagged. Fasttext and three algorithms of the results of the Convolutional Neural Network (CNN), BiLSTM, Logistic Regression, and Naïve Bayes classifier models have been compared. As a significant result, this paper obtained 0.996 AUC and 0.956 F-score using FastText and CNN.

In [43], Iranian user's comments about movies and cinema news have been collected on two websites, caffecinema.com and cinematicket.org, and have been tagged by three Persian-speaking persons using The Persent dictionary. Stacked-bidirectional-LSTM neural networks have been used for classification, using bag of words method and PCA algorithm to extract features, which has reduced the data to 200 dimensions.

In [44], authors collect, label and create a dataset of Persian-English code-mixed tweets. Then proceed to introduce a model which uses BERT pretrained embeddings as well as Yandex and dictionary-based translation models to automatically learn the polarity scores of these Tweets. This model outperforms the baseline models that use Naïve Bayes and Random Forest methods.

In [45], ArmanEmo, a human-labeled emotion dataset of more than 7000 Persian sentences labeled for seven categories has introduced. The dataset has been collected from different resources, including Twitter, Instagram, and Digikala. Labels are based on Ekman's six basic emotions (Anger, Fear, Happiness, Hatred, Sadness, Wonder) and another category (Other) to consider any other emotion not included in Ekman's model. Best model is XML-RoBERTa-large that achieves a macro-averaged F1 score of 75.39 percent across test dataset.

In [46], a dataset that includes the emotional sentences of Persian literary texts was created. The FastText and the Bi-LSTM model has compared for binary and multiclass classification. Then XLM-R based on a deep bidirectional transformer to extract features from text and the Catboost algorithm to guide the model to focus on the most relevant class has employed. This model can be used in all languages, especially low-resource languages and imbalanced datasets. In using this method, the accuracy of multi-class classification has been 52%. XLM-R model used for increasing accuracy. Then Balance Bagging Classifier SVM and k-fold learning were used to reduce class imbalances and Catboost algorithm was used. The accuracy in multi-class classification has increased to 72%. In binary classification, accuracy led to 82%.

## 5 Discussion and comparison

Previous sections explained the review method of the selected studies for sentiment analysis in Persian language. In this section, all 40 researches between 2018 and 2022 with the approaches and innovations of each are listed below. In each article describes the type of classification, method, accuracy, and dataset. Furthermore, some analytical reports regarding the planed analytical questions in Section 3 were presented as follows:

- AQ1: Which methods in machine learning are considered for Sentiment analysis in Persian language?
- AQ2: Which classifiers are categorized in Sentiment Analysis in Persian language?
- AQ3: What datasets are used for evaluating Sentiment analysis in Persian language?
- AQ4: What are the evaluation factors usually applied in Sentiment analysis in Persian language?

- AQ5: What are the innovations used for increasing accuracy in Sentiment analysis in Persian language?

Table 1. presents a comparison of sentiment analysis methods from 2018 and 2022 answer above questions.

| Reference | Objective | Classification type | Method | Accuracy | Dataset |
|---|---|---|---|---|---|
| [7] 2019 | A Hybrid Persian Sentiment Analysis Framework: Integrating Dependency Grammar Based Rules and Deep Neural Networks | Two-Class Classifier | Deep Learning (LSTM) + Dependency Grammar Based Rules | Digikala = 81%, Hellokish = 86 % | Digikala + Hellokish datasets |
| [8] 2019 | Opinion mining in Persian language using a hybrid feature extraction approach based on convolutional neural network | Two-Class Classifier, Five-Class Classifier | Feature Extraction = CNN, Classification = Bi-LSTM | Two-Class = 95 % Five-Class = 92 % | Digikala dataset |
| [9] 2019 | A novel approach to sentiment analysis in Persian using discourse and external semantic information | Three-Class Classifier | Group classification (MLP + Logistic + SMO) | 81 % | Persian hotel reviews dataset |
| [10] 2020 | DeepSentiPers: Novel Deep Learning Models Trained Over Proposed Augmented Persian Sentiment Corpus | Two-Class Classifier, Five-Class Classifier | Adding sentences to the dataset using the placement of synonymous words + CNN + Bi-LSTM | Two-Class = 91.98 %, Five-Class = 69 % | Digikala dataset |
| [11] 2019 | SentiFars: A Persian Polarity Lexicon for Sentiment Analysis | Three-Class Classifier | Translate several other languages into Persian to compose and build an emotional Persian dictionary + Logistic Algorithm | 95.92 % | Persian hotel reviews dataset |
| [12] 2021 | An Ensemble based Classification Approach for Persian Sentiment Analysis | Two-Class Classifier | Group classification (SVM + MLP + CNN) | 79 % | Persian hotel reviews dataset |
| [13] 2018 | Exploiting Deep Learning for Persian Sentiment Analysis | Two-Class Classifier | Autoencoder Neural Networks + CNN | 82 % | Comments of Iranian users on the topics of |

| Ref | Title | Classifier | Method | Accuracy | Dataset |
|---|---|---|---|---|---|
| | | | | | the film between 2014 - 2016 |
| [14] 2020 | Deep Persian sentiment analysis: Cross-lingual training for low-resource languages | Two-Class Classifier | Transitional learning + combine (LSTM + CNN) | 79 % | Digikala dataset |
| [15] 2020 | The application of Deep Learning in Persian Documents Sentiment Analysis | Three-Class Classifier | Feature Extraction = SCL Algorithm + CNN Neural Networks | 74 % | SentiPers (User comments about digital products) |
| [16] 2020 | Tehran Stock Exchange Prediction Using Sentiment Analysis of Online Textual Opinions | Two-Class Classifier | Manually creating a dictionary in the field of stock exchange to predict the Iranian stock market + bagging j48 Algorithm | 85 % | Iranian stock market |
| [17] 2020 | EXPLOITING BERT TO IMPROVE ASPECT-BASED SENTIMENT ANALYSIS PERFORMANCE ON PERSIAN LANGUAGE | Three-Class Classifier | Aspect-level analysis + BERT Algorithm | 91 % | Psrs-ABSA4 dataset |
| [18] 2020 | Optimizing Annotation Effort Using Active Learning Strategies: A Sentiment Analysis Case Study in Persian | Three-Class Classifier | Active learning + combine (LSTM + CNN) + Transfer data to Latent space | F1-Score = 80 % | MirasOpinion (User comments on Digikala site) |
| [19] 2019 | A COMBINED DEEP LEARNING MODEL FOR PERSIAN SENTIMENT ANALYSIS | Three-Class Classifier | Extract local feature with CNN + LSTM for Classification | F-Score = 85 % | Dataset1 = 9066 Comments, Dataset2 = 2550 Comments in Social Media |
| [20] 2019 | Pars-ABSA: an Aspect-based Sentiment Analysis dataset for Persian | Three-Class Classifier | Aspect-level analysis + Bi-LSTM | 85 % | Pars-ABSA Dataset |

| Ref / Year | Title | Classifier | Method | Accuracy | Dataset |
|---|---|---|---|---|---|
| [21] 2018 | Words Are Important: Improving Sentiment Analysis in the Persian Language by Lexicon Refining | Five-Class Classifier | Machine learning + Lexicon | 45 % | PerView (16000 User Comments Digikala) |
| [22] 2020 | HOMPer: A new hybrid system for opinion mining in the Persian language | Two-Class Classifier, Five-Class Classifier | A hybrid method for ranking comments using machine learning + feature selection + normalization | Two-Class = 82 %, Five-Class = 62 % | 16,000 Persian customers' review |
| [23] 2020 | Unsupervised aspect-based Sentiment Analysis in the Persian language: Extracting and clustering aspects | Three-Class Classifier | Aspect-level analysis + K-means Clustering | 86 % | SentiPers |
| [24] 2019 | LexiPers: An ontology-based sentiment lexicon for Persian | Three-Class Classifier | Sentiment lexicon + create an ontology + KNN Algorithm | 66 % | LexiPers |
| [25] 2021 | SentiPers: A Sentiment Analysis Corpus for Persian | Three-Class Classifier | | 67 % | SentiPers |
| [26] 2019 | An improved evidence-based aggregation method for sentiment analysis | Five-Class Classifier | Adding sentences to the database using synonyms words + Use a ranking system + Glossary | 80 % | Hotel dataset + User comments about the restaurant |
| [27] 2018 | Sentence-level sentiment analysis in Persian | Three-Class Classifier | tokenization, normalization and stop words removal + SVM, MLP and Naïve Bayes classifier | 89.62 % | PerSent lexicon dataset + The Persian movie review dataset |
| [28] 2021 | Adversarial Weakly Supervised Domain Adaptation for Few Shot Sentiment Analysis | Two-Class Classifier | ParsBERT + domain classifier + sentiment classifier | 87 % formal 77 % colloquial f1-score = 80 % formal 69 % colloquial | Digikala and Sentipers |

| Ref | Title | Classifier | Method | Accuracy | Dataset |
|---|---|---|---|---|---|
| [29] 2021 | The Impact of Active Learning Algorithm on a Cross-lingual model in a Persian Sentiment Task | Two-Class Classifier | RNN + GRU + MUSE and VecMap | 90.51 % | Amazon Food Reviews + Persian Snapp Food reviews |
| [30] 2021 | ParsBERT Post-Training for Sentiment Analysis of Tweets Concerning Stock Market | Three-Class Classifier | ParsBERT | Precision = 82.04 %, Recall = 82.6 %, F-score = 81.83 % | Twitter |
| [31] 2021 | Sentiment Analysis of Persian Instagram Post: a Multimodal Deep Learning Approach | Three-Class Classifier | bidirectional GRU + two-dimensional convolution | 92%, f1-score = 92 % | MPerInst |
| [32] 2020 | The effect of aggregation methods on sentiment classification in Persian reviews | Five-Class Classifier | Cross-ratio aggregation + Naïve Bayes and SVM classifier | 81.7 % Precision = 67.6 %, Recall = 69.2 %, F-score = 68.4 % | Digikala 2017 |
| [33] 2022 | Twitter sentiment analysis from Iran about COVID 19 vaccine, Diabetes & Metabolic Syndrome: Clinical Research & Reviews | Three-Class Classifier | CNN-LSTM Hybrid Model | 81.77 % | Foreign-Vaccine + Homegrown-Vaccine dataset |
| [34] 2020 | Sentiment Analysis of Informal Persian Texts Using Embedding Informal words and Attention-Based LSTM Network | Two-Class Classifier + Three- Class Classifier | Use of Attention-based LSTM Network + Sentiment Analysis of Informal Persian Texts Using fasttext Embedding | 80.3 % | Filimo Dataset + Taaghche Dataset |
| [35] 2020 | Producing An Instagram Dataset for Persian Language Sentiment Analysis Using Crowdsourcing Method | Three- Class Classifier | Introducing the first Persian Instagram comments dataset + Use the skip-gram word2vec model to validate the insta-Text dataset. | --- | Insta-Text |

| Ref/Year | Title | Classifier | Method | Accuracy | Dataset |
|---|---|---|---|---|---|
| [36] 2019 | PerSent 2.0: Persian Sentiment Lexicon Enriched with Domain-Specific Words | Two-Class Classifier | SVM + Naïve Bayes | 81.3 % | PerSent 2.0 |
| [37] 2020 | Persian Sentiment Analysis without Training Data Using Cross-Lingual Word Embeddings | Two-Class Classifier | Using VecMap method to make English and Persian word embeddings. | F1-score = 76.16 % | Snapp food reviews + Amazon Fine Food Reviews + tailored English |
| [38] 2021 | Deep Learning-based Sentiment Analysis in Persian Language | Three-Class Classifier | CharEmbed + Word2Vec + LSTM | F1-score = 78.3 % | Digikala |
| [39] 2021 | Jointly Modeling Aspect and Polarity for Aspect-based Sentiment Analysis in Persian Reviews | Two-Class Classifier | Identify aspects and determine polarity simultaneously + Group Classification (CNN, LSTM, Bi-LSTM, GRU) | 80 % | Cinema Ticket |
| [40] 2021 | Adopting Transition Point Technique for Persian Sentiment Analysis | Two-Class Classifier | proposed a framework using TF-IDF and transition point + SVM, Naive Bayes, MLP and CNN for classification | 97 % | Persian movie reviews Dataset (cinema ticket + caffe cinema) |
| [41] 2022 | Extending Persian sentiment lexicon with idiomatic expressions for sentiment analysis | Two-Class Classifier | CNN + Extending Persian sentiment lexicon with idiomatic expressions + introduction of PerSent 2 | 78.8 % | PerSent 2 |
| [42] 2021 | Persian sentiment analysis of an online store independent of pre-processing using convolutional neural network with fastText embeddings | Two-Class Classifier | CNN + using FastText Embedding + pseudo-labeling to tag unlabeled data | 99.6 % F-Score = 95.6 % | Digikala |
| [43] 2021 | Sentiment Analysis of Persian Movie Reviews Using Deep Learning | Two-Class Classifier | stacked-bidirectional-LSTM + PCA and Bag of | 95.61 % | Persian Movie reviews |

| Paper | Title | Classifier | Method | Performance | Dataset |
|---|---|---|---|---|---|
| | | | Words for feature extraction | | |
| [44] 2021 | Sentiment Analysis of Persian-English Code-mixed Texts | Three-Class Classifier | Bi-LSTM + Attention + BERT Word Embedding | 66.17 % F-score = 63.66 % | Persian-English Code-mixed Texts |
| [45] 2022 | ARMANEMO: A PERSIAN DATASET FOR TEXT-BASED EMOTION DETECTION | Seven-Class Classifier | ARMANEMO + Zero-shot tests using XLM-Roberta-large | F1=75.3 % | ARMANEMO |
| [46] 2022 | Deep Emotion Detection Sentiment Analysis of Persian Literary Text | Two-Class Classifier | XLM-R, Catboost Gradient, Decision Tree | 82 % | JAMFA corpus |

## 5.1 Datasets

One of the challenges of sentiment analysis in Persian Language is the lack of large labeled dataset. To overcome this problem and increase the size of the dataset, data enhancement and text reinforcement techniques have been used [7, 10]. In addition, in some studies, due to the lack of datasets in a particular field, the authors were forced to build crawler robots that extracted the desired information from various websites to build their own personal datasets [16]. In many researches, a larger dataset has been created by combining several datasets so that machine learning models can be trained with more data [7, 13, 19, 26]. Of course, using methods such as translating several different languages into Persian, Transitional learning, and Active learning, a relatively large dataset has also been analyzed in Persian [8, 11, 14, 18]. Table 2 describes all the datasets used in the researches between 2018 and 2022 with brief descriptions.

Table 2. Datasets with details

| Paper | dataset | description of the dataset |
|---|---|---|
| [7] | Digikala and Hellokish | Digikala is one of the best startup and online shop. This dataset contains 1500 positive comments and also 1500 negative comments. Another dataset is Hellokish that this dataset consists of 1800 positive and 1800 negative hotel reviews. |
| [8] | Digikala | This dataset from Digikala has two and five classes. class 1: "very bad" contains 7,521 comments. class 2: "bad" contains 10,227 comments. class 3: "Moderate" contains 102,439 comments. class 4: "Good" contains 16,040 comments. class 5: "Excellent" contains 15,002 comments. |
| [9] | Persian hotel reviews | The dataset is collected from Persian hotel reviews that contains 761 documents and 3613 sentences included in these documents. The distribution of different classes in sentences and documents are respectively (neg, obj, pos) = (27%, 13%, 60%) and (neg, obj, pos) = (22%, 7%, 71%). |

| [10] | Digikala | Another Digikala dataset is used in this paper that has five classes. class 1:" Furious" contains 40 sentences. class 2: "Angry" contains 697 sentences. class 3: "Neutral" contains 3,152 sentences. class 4: "Happy" contains 2,184 sentences. class 5: "Delighted" contains 1,342 sentences.<br><br>Another type of this dataset is two classes that were used in this research. class 1: "Negative" contains 737 sentences. class 2: "Positive" contains 3,526 sentences. |
|---|---|---|
| [11] | SentiFars | Four lexicons have been used to create FarsNet which all of them have three labels (Negative. Neutral and Positive). 1- SenticNet 3.0: 14,227 entries. 2- Liu's Polarity Lexicon: 6784 entries. 3- NRC Emotion Lexicon: 14182 entries. 4- SentiWordNet 3.0: 117659 entries. |
| [12] | Persian hotel reviews | This dataset consists of 3000 reviews: 1500 positive and 1500 negative comments about the hotel. |
| [13] | Comments of users | This dataset contains comments of Iranian users on the topics of the film between 2014 - 2016. There are two types of labels in the dataset: positive or negative. The reviews were manually annotated by three native Persian speakers aged between 30 and 50 years old. This dataset contains 10,000 comments (Negative:5000, Positive:5000). |
| [14] | Digikala and Amazon | The datasets used in this article include the Amazon database which contains 158,045,100 comments classified in five categories and the other is the Digikal database which contains 305,639 comments classified in two categories. |
| [15, 23, 25,28] | SentiPers | This corpus contains more than 26,000 Persian sentences of user's opinions from the digital product domain and benefits from special characteristics such as quantifying the positiveness or negativity of an opinion through assigning a number within a specific range to any given sentence. Each appliance in SentiPers contains a set of characteristics and sentences. The content of sentences is user's opinion about electronic goods. The sentences are classified in positive, negative and neutral classes. A number in the interval [-2, 0, +2] is assigned to each opinion that presents the score of the sentence in positive or negative classes. |
| [16] | Iranian stock market | Since the dictionaries available for Persian language are not for analyzing emotions in the field of stock exchanges, the authors have created a special dictionary for this field. Text comments have been collected from www.sahamyab.com/stocktwits, which is the only social network for the stock market in Iran. This paper used the unbalanced dataset for making the lexicon. This dataset included 2,125 bearish and 6,248 bullish comments. |
| [17] | Psrs-ABSA4 | The Psrs-ABSA4 dataset includes user feedback on products and services. This dataset contains 5,114 targets with positive polarity, 3,061 with negative polarity and 1,827 targets with neutral polarity. Total number of comments is 5,602 which shows that many comments have more than one aspect. |
| [18] | MirasOpinion | This database includes user comments on the Digikala site. This is a huge database and almost 93,000 unlabeled comments have been extracted from it. |

| | | These comments are then labeled by many people in three classes: positive, negative, and neutral. positive comments: 49,515. negative comments: 14,882. neutral comments: 29,471. |
|---|---|---|
| [19] | PE dataset and PP dataset | PE dataset: To create this dataset, a crawler was used to collect data from the Digikala site for electronic products. The name of this dataset is "PE" which stands for Persian Electronic dataset. There are three classes with a total of 9,066 comments. class 1: "positive", class 2: "negative" and class 3: "neutral" that each of them contains 3,022 comments.<br>PP dataset: Also, the twitter API has been used to create a second dataset on the political issue in Persian. The name of this dataset is "PP", which stands for Persian Political dataset. There are three classes with a total of 2,550 comments. class 1: "positive", class 2: "negative" and class 3: "neutral" that each of them contains 820 comments. |
| [20] | Pars-ABSA | Pars-ABSA was created from collected user reviews from Digikala site. There are 3 classes with positive, negative and neutral labels. Number of targets: 10,002, number of targets with positive polarity: 5,114, number of targets with negative polarity: 3,061, number of targets with neutral polarity: 1,827. |
| [21] | PerView | In this paper, two new sources for solving the problem of lack of Persian resources in the field of emotion analysis are introduced, including a glossary carefully labeled with emotional words (PerLex) and a new handmade dataset with about 16,000 rated documents (PerView). The PerView comments were collected from July 2016 to February 2017. It contains customer's comments about digital equipment including cell phones, cameras, and computer peripherals. |
| [22] | Persian customers review | This dataset contains 16,000 Persian customer's reviews. |
| [24] | LexiPers | In this paper generated a sentiment lexicon based on the FarsNet ontology. FarsNet, a publicly available ontology in Persian containing more than 20,000 synsets. There are three classes in LexiPers (negative, netural, positive) and the number of all Synsets is 20432. Negative class: 3,306, neutral class: 14,281 and positive class: 2,845. |
| [26] | Hotel review + User comments about the restaurant + SentiStrength Lexicon | Hotel dataset: This dataset collected from TripAdvisor and contains customers' reviews about different hotels all over the world and includes 3000 reviews. In this dataset, one- to five-star reviews are distributed equally and 600 reviews have been assigned to each class.<br>Restaurant dataset: This dataset has been extracted from NY CitySearch website and contains customers' reviews about restaurants. It includes 3000 reviews distributed equally in the five existing classes. In this dataset, 600 reviews have been assigned to each class. |
| [27] | PerSent lexicon and The Persian movie review | The lexicon contains 1500 Persian words along with their part-of-speech tag and their polarity. The movie reviews contain 1000 positive and 1000 negative sentences. The movie has been collected from |

| | | www.caffecinema.com and www.cinematicket.org, the collected movie reviews are from 2014-2017 movies. |
|---|---|---|
| [28] | Digikala and Sentipers | Dataset of formal and colloquial sentences include 20,000 sentences taken from expert review texts and 20,000 sentences taken from user comments at the bottom of products. It has hypothesized that the first (expert reviews) is in the formal domain and the second (user comments) is in the colloquial domain. The sentences of this dataset do not contain any sentiment tags. |
| [29] | Amazon Food Reviews and Persian Snapp Food reviews | Snapp Food dataset is contained of 6,456 reviews both negative and positive. And Amazon dataset, have 90,000 reviews out of 104,536 as the train dataset and 14,536 reviews as the test dataset. |
| [30] | Twitter | over 12000 Persian tweets including the stock market keyword have been crawled from twitter. They are labeled manually in three different categories of positive, neutral and negative. |
| [31] | MPerInst | manually downloaded 512 public Instagram image-text pairs and save them in an xlsx format. The average length of the texts in the dataset is 14.51 words. |
| [32] | Digikala 2017 | It contains of 16000 reviews of four datasets of customer reviews about different brands of cell phones (Apple-LG, Huawei-Sony, Samsung, note 5) were extracted from Digikala.com in the period of July 2016 and February 2017. |
| [33] | Foreign-Vaccine dataset and Homegrown-Vaccine dataset | The search keywords for the homegrown vaccine were "واکسن برکت، واکسن داخلی" (Barekat vaccine, homegrown vaccine). Accordingly, the keywords for retrieving tweets about foreign vaccines were "واکسن آسترازنکا، واکسن آکسفورد، واکسن فایزر، واکسن مدرنا، واکسن سینوفارم، واکسن خارجی" (AstraZeneca vaccine, Oxford vaccine, Pfizer vaccine, Moderna vaccine, Sinopharm, foreign vaccine). all Persian-language tweets posted in the time frame from April 1, 2021 to 30 September 2021 that related to the keywords were retrieved. The Homegrown-Vaccine dataset contained 400839 (49.9%) tweets, and the Foreign-Vaccine dataset included 402439 (50.09%) tweets accordingly. |
| [34] | Filimo and Taaghche | 15678 comments were extracted from the Filimo website. These two datasets contain user's opinions about movies and series. Filimo database contains 4712 comments which are classified into three categories: Negative: 995, Neutral: 828 and Positive: 2889.<br>On the "Taaghche" website, users rate books using numbers 1 to 5. Taaghche database contains 2360 views, which are classified into two categories: Negative Comments: 986 and Positive Comments: 1320 |
| [35] | Insta-Text | This dataset contains 8,512 comments from various people under the HALA_KHORSHID page posts, which were collected with the Instagram-scraper tool and tagged by 20 computer engineering students in three classes: positive comments: 2,780, neutral comments: 2,495 and negative comments: 3,237. |
| [36, 41] | PerSent 2 | This sentiment lexicon is developed by PerSent 1, which has 1000 idiomatic expressions and 3000 words, which are labeled in two classes of positive and negative with a decimal accuracy of 0.1. -1 means the word with |

| | | the most negative charge and +1 means the word with the most positive charge. |
|---|---|---|
| [37] | Amazon Fine Food Reviews + Snapp food reviews + tailored English | The Snapp food database contains positive and negative comments from customers about food services, which include 3,096 positive comments and 2,998 negative comments, and has been used as a test database.<br>Amazon Fine Food Reviews and tailored English datasets have been used as training datasets. The tailored English database consists of 5 classes that are balanced into 2 positive and negative classes and a total of 104,536 comments. Positive Comments: 90,000 and Negative Comments: 14,536 Comments. |
| [38] | Digikala | This dataset has 62,738 user comments on the Digi Kala site, which are tagged in three categories. Positive comments: 36,482 and negative: 15,871 and neutral: 10,385 |
| [39] | CinemaTicket | This dataset contains user comments on movie ratings on the CinemaTicket website (www.cinematicket.org), which consists of 3,576 comments, which are classified into fourteen different categories and two classes, positive and negative. Positive Comments: 1,878 and Negative Comments: 1,698 Comments. |
| [40] | Persian movie reviews Dataset (cinematicket + caffecinema) | This dataset contains 1000 positive comments and 1000 negative comments about movie topics in Cinema Ticket sites (www.cinematicket.org) and Cinema Cafe (www.caffecinema.com). Comments were collected between 2014 and 2016. |
| [42] | DigiKala | This database contains 2,932,747 records that have been extracted from DigiKala and have been labeled in two categories, negative and positive. There are 1,800,000 positive comments, 400,000 negative comments and 700,000 indeterminate comments. Tags based on two options "I suggest" and "I do not suggest". |
| [43] | Persian Movie reviews | Iranian user's comments about movies and cinema news have been collected on two websites, caffecinema.com and cinematicket.org, and have been tagged by three Persian-speaking people between the ages of 30 and 50. 1021 positive comments and 989 negative comments were extracted, which led to a new dataset with 2010 comments. |
| [44] | Persian-English Code-mixed Texts | The dataset contains 3640 comments classified into three classes: negative, positive and neutral. Comments are collected from the Twitter social network. |
| [45] | ARMANEMO | This data collection includes opinions of users on Twitter, Instagram, Digikala and etc. The dataset contains more than 7000 comments classified into 7 classes. (Anger, fear, joy, hatred, sadness, surprise and others). |
| [46] | JAMFA corpus | The JAMFA dataset contains 2241 emotional sentences collected from Persian literary books, and the comments are labeled in two categories. 432 positive comments and 1809 negative comments. To create this dataset, novels by famous Iranian writers such as Simin Daneshvar, Jalal Al Ahmad and Samad Behrangi were used. |

## 5.2 Datasets along with the accuracy of the algorithms applied to them

Data is one of the most important elements in machine learning and the more data there is, the more complex algorithms can be applied on them and the higher the prediction accuracy. Of course, the amount of data is not directly related to the accuracy of the algorithms on them. Data alone are not a good parameter for analysis, problem-solving and innovative approaches are important. In this section, we have reviewed the articles by type of classification such as two-class, three-class and five-class.

Two-class classifier have two labels 0 or 1 (negative, positive) for each comment. An important result obtained in this review is that there is no direct relationship between the number of data and the accuracy of the system. For example, in [10], with increasing data, its accuracy decreases. But why does this occur? One of the reasons that [10, 39] is so accurate is that it uses Bidirectional Long Short Term Memory Neural Networks (Bi-LSTM). This model is one of the best Neural Networks used for issues that depend on place or time. Bi-LSTM is an improved LSTM and LSTM is a subset of RNNs that have improved the performance of RNNs. One of the problems with Recurrent Neural Networks is that they have a short memory, meaning they can't store information from previous layers or steps for long. To overcome this problem, use LSTMs and Bi-LSTMs that these two architectures can store information for a long time. Another reason that [10] has achieved high accuracy is that it uses the Data Augmentation technique. Using this technique, new data can be generated. That is, by replacing synonymous and contradictory words and generating new textual data, data diversity can be increased.

All three-class research have labeled in -1, 0, or 1 (negative, neutral, positive) for each comment. [20] has very little data volume but has reached good accuracy. One of the reasons that [20] has achieved such high accuracy is that it has used Bi-LSTM for sentiment analysis. As mentioned in the previous section, Bi-LSTM is a great tool for text processing and sentiment analysis. As a result, Transformers such as BERT can achieve higher accuracy in various NLP tasks [17], Bi-LSTM neural networks are also the best choice after BERT [20] and local feature extraction with CNN + LSTM and ParsBERT [30] for classification is a more reliable method [19]. Figure (7) shows that three-class datasets mostly used in Persian sentiment analysis approach papers between 2018 and 2022.

Finally, we review articles that have 5 output classes which among them, [8] achieved the best accuracy. In [26] sentences are added to the dataset using synonyms words and as well as using a ranking system and glossary. When the amount of data in the dataset is low, using this technique can increment accuracy. One of the advantages of NLP is that you can use text reinforcement or text augmentation. Using this technique, you can increase the amount of data and achieve higher accuracy. Among the papers in this section, [26] has the smallest dataset which by using this technique, has increased the volume of data and has reached the highest accuracy. As a result, the use of text reinforcement or text augmentation technique [26] can greatly increase the accuracy of prediction and is also recommended for small datasets.

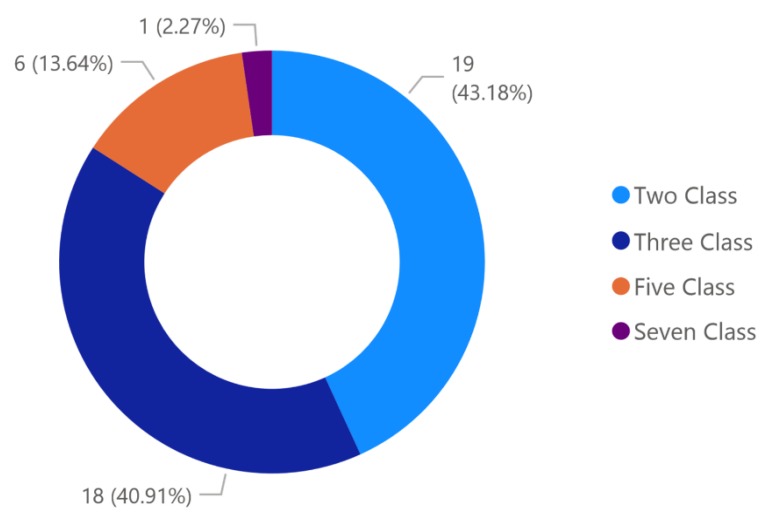

**Fig. 7.** Distribution of datasets by class numbers.

Figure (8) shows main approaches that been used in sentiment analysis in Persian language papers between 2018 and 2022 and the average accuracy of each algorithm.

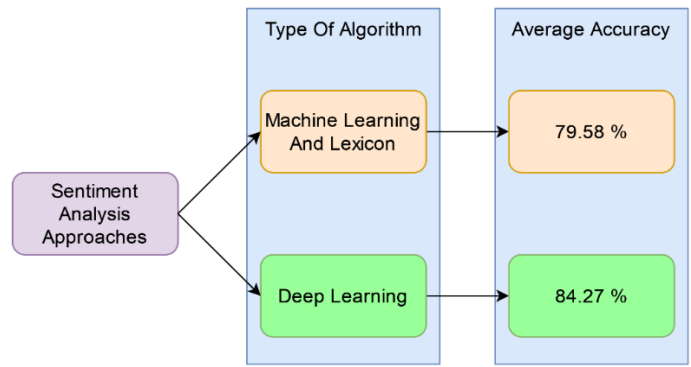

**Fig. 8.** average accuracy of main approaches.

## 5.3 Challenges, Open issues and trends

Due to applying the SLR process on the study assortment of sentiment analysis in Persian language, the following research challenges as the open issues are presented as the AQ6.

- AQ6: What are the future researches directions and open perspectives of Sentiment analysis in Persian?

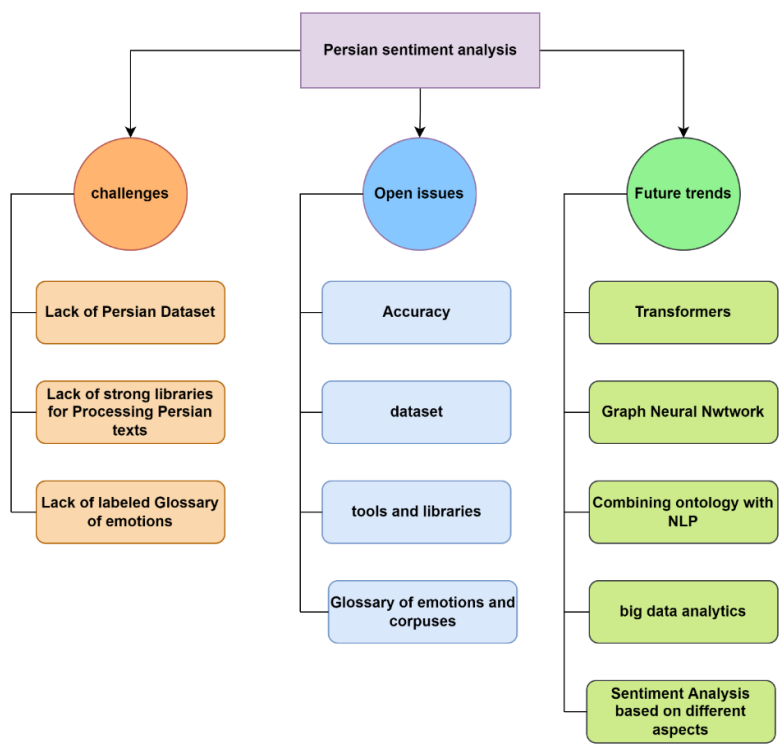

**Fig. 9.** challenges, open issues and trends in this review.

## 6 Conclusion

In this paper, 40 researches between 2018 and 2022 have been collected and for each, algorithms, approaches and dataset have been explained, reviewed and analyzed which among them, transformers such as BERT have achieved higher accuracy in Sentiment Analysis. LSTM, Bi-LSTM neural networks and CNN are also the best choice after BERT. In addition, different techniques can increase the accuracy of prediction in Sentiment Analysis, such as Integrating Dependency Grammar Based Rules, Data Augmentation technique (adding sentences to the dataset using the placement of synonymous words), translating several languages into Persian by writing and building an emotional Persian dictionary, adding sentences to the dataset using synonyms words.